\documentclass{article}

\usepackage{arxiv}

\usepackage[utf8]{inputenc} 
\usepackage[T1]{fontenc}    
\usepackage{hyperref}       
\usepackage{url}            
\usepackage{booktabs}       
\usepackage{amsfonts}       
\usepackage{nicefrac}       
\usepackage{microtype}      
\usepackage{lipsum}

\usepackage{subfigure}

\usepackage{amsmath,amsfonts,bm}

\def\eqref#1{equation~\ref{#1}}

\def\1{\bm{1}}

\DeclareMathAlphabet{\mathsfit}{\encodingdefault}{\sfdefault}{m}{sl}
\SetMathAlphabet{\mathsfit}{bold}{\encodingdefault}{\sfdefault}{bx}{n}

\usepackage{amsmath}
\usepackage[mathscr]{euscript}
\usepackage{amssymb}
\usepackage{tabularx}
\usepackage{textcomp}
\usepackage{gensymb}
\usepackage{multirow}
\usepackage{graphicx}
\usepackage{appendix}
\usepackage{csquotes}

\usepackage{float}
\usepackage{algorithm}
\usepackage[noend]{algpseudocode}

\usepackage{ntheorem}

\usepackage[english]{babel}

\DeclareUnicodeCharacter{0301}{\'{e}}

\usepackage[
backend=biber,
style=alphabetic,
sorting=ynt
]{biblatex}
\usepackage{titling}
\usepackage{listings}
\usepackage{xcolor}

\usepackage{authblk}
\usepackage[document]{ragged2e}
\definecolor{dkgreen}{rgb}{0,0.6,0}
\definecolor{gray}{rgb}{0.5,0.5,0.5}
\definecolor{mauve}{rgb}{0.58,0,0.82}

\usepackage{xcolor}

\title{Cogment: Open Source Framework For Distributed Multi-actor Training, Deployment \& Operations}

\author{AI Redefined}
\affil{\texttt{contact@ai-r.com}}
\author{Sai Krishna Gottipati}
\author{Sagar Kurandwad}
\author{Clodéric Mars}
\author{Gregory Szriftgiser}
\author{François Chabot}
\affil{\texttt{\{first\_name\}@ai-r.com}}

\begin{document}
\maketitle

\begin{abstract}

Involving humans directly for the benefit of AI agents' training is getting traction thanks to several advances in reinforcement learning and human-in-the-loop learning. Humans can provide rewards to the agent, demonstrate tasks, design a curriculum, or act in the environment, but these benefits also come with architectural, functional design and engineering complexities. We present Cogment, a unifying open-source framework that introduces an actor formalism to support a variety of humans-agents collaboration typologies and training approaches. It is also scalable out of the box thanks to a distributed micro service architecture, and offers solutions to the aforementioned complexities.

\end{abstract}


\section{Introduction}

When systems start to be too complex or too large for traditional methods to produce quality results in a reasonable time frame, AI is often brought in. But many systems are too critical to be fully trusted to AI agents (e.g. medical applications). Thus human and AI agent collaboration becomes a central aspect of such systems.  This requires, however, an environment in which humans and AIs can operate and train together. Humans leveraging AI's insights is widespread, in the field of Business Intelligence for example \cite{chen2012business}. Humans learning from AIs is also relatively common, for example in aircraft pilot simulation training \cite{jones1999automated}. In the past few years, human involvement has grown beyond data annotation to become what we call human-in-the-loop training; providing AIs with feedback and guidance. In many domains, the performance of AIs was shown to be enhanced by taking feedback from humans \cite{zhang_leveraging_2019}, or simply by learning from human demonstrations \cite{IL-survey}. However, there was no unifying framework enabling researchers to quickly develop application supporting human-in-the-loop learning, nor for engineers to deploy those at scale. Cogment is designed to meet those needs. It's a framework facilitating the development and deployment of projects involving multiple actors, humans or AI agents, interacting with each other in an environment, simulated or not.  

In this document, we will explore the main features of Cogment and why they make human-in-the-loop and, more generally, multi-actor architectures easier to implement and operate. We will then give an overview of the application domains for Cogment before diving deeper into how Cogment is designed and works. Finally we will go through a few concrete use cases built with Cogment.

\section{Key Features}

\begin{figure}
\centering
\includegraphics[width=0.8\columnwidth]{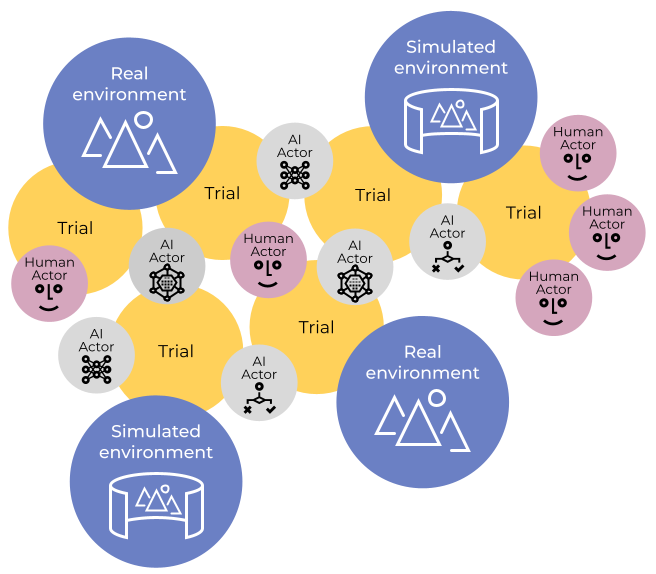}
\caption{Cogment orchestrates the running of \emph{trials} involving AI and human \emph{actors} in simulated or "real-world" \emph{environments}. Each trial involves one or multiple actors and runs in one environment. Multiple trials can run concurrently, and they can share actors or environments.}
\label{fig:trials}
\end{figure}

\subsection{Multi-actor framework}

Cogment provides a multi-actor framework where multiple heterogeneous actors interact with an environment during trials. In Cogment, actors can be either humans or AI agents of any kind: learning or static, using machine learning models or not. Cogment thus enables AI agents and humans to interact and train together in shared environments, in any configuration; an actor can interact in one or several trials (see Figure \ref{fig:trials}).

One key capability Cogment offers, which supports many use cases, is the training of AI agents using the inputs from the humans present in the same environment. Compared to machines, humans have a limited work capacity and can't operate at the same speeds: this makes their presence scarce in such environments, in terms of data contribution. It is therefore desirable to leverage all the possible inputs that can be generated by them. This includes evaluative feedback, i.e. humans evaluating the performance of AIs at a given task, and demonstrations, i.e. humans performing a task for the benefit of an AI \cite{zhang_leveraging_2019}. Cogment supports both these types (among others) of human-in-the-loop learning. 
When it comes to evaluative feedback, it can be desirable to mix several sources of such evaluation: multiple humans evaluating the same actions of an AI or mixing the subjective feedback of a human with an objective performance measurement. That's why Cogment supports \textbf{rewards from multiple sources} and aggregates them.

Another aspect of human-provided evaluative feedback is an inherent lag between a feedback-calling action, its perception by the evaluating human, and its subsequent evaluation. To facilitate that, Cogment supports \textbf{retroactive rewards}, rewards that apply to a past action, while maintaining online learning capabilities.

One foundational aspect of Cogment is the definition of contracts for the interactions between the actors and the environment, in the form of an action space and an observation space. These definitions make it easy to decouple the development process of each actor and the environment implementation while also enabling a very interesting feature: \textbf{implementation swapping}.

While observation and action spaces define the interface and the way actors can interact with the environment, their \emph{implementation} defines how they behave in accordance with this interface. Several implementations can share the same interface, and if they do, they can be swapped. This opens the door to several training and operating topologies that are especially relevant when involving humans.

Let's say we are training two AI agents in an environment with two humans. The problem with starting the training with this setup is that humans will interact with \emph{very} dumb AI agents for a while before these start to do something useful, which is not an efficient use of humans. Cogment's implementation swapping allows more interesting training setups to work around this issue.

\begin{itemize}
    \item \textbf{Bootstrap with pseudo-humans:} implement simple rule-based AI agents simulating or mimicking the behavior of humans, run a lot of fully simulated trials with this setup. Once the AI agents have reached a good performance level, start involving the actual humans.
    \item \textbf{Bootstrap with business expertise based AI agents:} implement the two agents using, for example, a rule-based system to provide some value to the human, and add some stochasticity to bolster variety. These agents will start generating data that can be used to train Machine Learning (ML) based policies. Once these are good enough, the ML based implementations can replace the initial ones.
\end{itemize}

These two are simple examples of training curriculums that can easily be implemented using Cogment.

While the discussion so far has been focused on making the most out of human resources via efficient human-in-the-loop learning setups, these capabilities can also be extended to situations where there's only different AI agents interacting with each other. Implementation swapping can be used to create complex training and evaluation scenarios. For example, training several AI agents in self play trials, before moving to mixing and matching AIs in another set of trials.

\subsection{Operationalized framework}

When it comes to training AIs in simulations or real world systems, and even more when including humans as evaluators or demonstrators, the required software involves a \textbf{lot of moving pieces}. Simulation engines can be as simple as applying the rules of a board game \cite{alphazero} or as complex as simulating the airflow in a building \cite{autodesk_cfd}. They use a wide range of technologies and architectures. Furthermore, "real-world" sensors and actuators systems come with a variety of technical properties and requirements. While the Machine Learning community is mostly using Python-based libraries such as PyTorch \cite{pytorch}, TensorFlow \cite{tensorflow} or scikit-learn \cite{scikit}, the larger AI community uses a variety of software platforms such as planners \cite{pomdp_solve} or constraints solvers \cite{gecode}. For humans to interact with software, a Graphical User Interface (GUI) is usually warranted. Popular ways of developing GUIs include the use of web technologies \cite{react,angular}, or 3D, real-time video game engines \cite{unity,unreal}.

Another challenge is the need to generate a large quantity of data for the training of AIs. This usually requires the ability to run a large number of \textbf{trials} (typically referred to as episodes or trajectories in the reinforcement learning literature). In some instances, a lot of AI agents might be required to join these trials. Yet another added complexity is the need for a lot of humans to interact with the different trials, (e.g, crowd sourcing). These humans might also be participating from different locations, requiring remote connecting to the simulation instances. 

Cogment has been designed to address these issues for its users. From the get go, \textbf{Cogment applications are distributed}, every part being implemented as a micro service communicating with the others using a technology-agnostic protocol; for further details see Section \ref{sec:architecture}. With Cogment, a developer, researcher or data scientist is able to start working on their personal computer and, ultimately, have their work being deployed on a high performance cluster to support a larger load without the need to break the existing implementation, what is called \textbf{integration discontinuity} \cite{muratori_2014}. In the same fashion, a Cogment application might begin with integrating a simulation before moving to its real-world counterpart, in order to facilitate the sim-to-real process and to continuously operate and train the AIs. Cogment \textbf{decouples the hosting and distribution concerns from the AI design, implementation and evaluation}.

\section{Applications}

The multi actor framework provided by Cogment lends itself to a wide variety of applications. In this section we will take a look at state of the art methods Cogment supports and facilitates.

\subsection{Reinforcement Learning}

Reinforcement Learning has already seen a lot of success in board games \cite{tdgammon,expertiteration,alphazero} and video games \cite{atari,alphastarblog,dota}. More recently, RL is being used in real world applications like manipulation tasks \cite{openai2021asymmetric}, active localization \cite{dal,anl}, autonomous navigation of stratospheric balloons \cite{stratoshpere}, de novo drug design \cite{pgfs,maximum2020}, autonomous driving \cite{wayverl} and many others. For computationally efficient training of these RL algorithms, one needs to be able to launch multiple trials simultaneously and gather diverse experiences from the environment. Cogment supports this. It can launch multiple trials, synchronously or asynchronously, using multiple agents or multiple instances of the same agent. For sample efficient learning of algorithms, one has to learn from the old samples - Cogment supports this as well by storing all the relevant information in log files in addition to the standard usage of a replay buffer. 

STEAM \cite{steam} is one of the first multi-actor frameworks that demonstrated the importance for unifying multiple teamwork capabilities like joint intentions \cite{jointintentions-Cohen1991}, \cite{jointintentions-levesque} and shared plan theories \cite{sharedplan1}, \cite{sharedplan2}. Another framework, ISAAC \cite{isaac} was introduced to help humans analyze, evaluate, and understand team behaviors in an offline setting. More recently, there has been significant progress in cooperative multi-agent games. A prominent one among those is the game of Hanabi, which is a very challenging partial-information cooperative multi-agent game. While the current research \cite{hanabiactiondecoder,hanabiactiondecoderog,hanabigoogle,hanabiog,hanabisarath,hanabitheoryofmind} has been focused on centralized training of agents, Cogment provides a way to train multiple agents in a distributed and decentralized manner, thus increasing its computational efficiency and enabling faster progress in multi-agent reinforcement learning research. These features of Cogment could also be used for learning emergent behaviours in a multi-agent competitive and cooperative setting \cite{emergent}. 

In some cases, an agent performs better if it takes input from, or is bootstrapped by another agent. For example, in expert-iteration \cite{expertiteration} based approaches like \cite{alphazero}, a policy that imitates the Monte Carlo Tree Search (MCTS) agent is learnt. The MCTS agent in turn initializes its prior probabilities based on the policy network's output. More generally, RL can be combined with other learning or non-learning based algorithms; for example, genetic algorithms \cite{rl-genetic-gated,rl-genetic-taylor} or heuristic search \cite{rl-heuristic-search}. Cogment provides an efficient way to run such algorithms via the use of actor implementations. In offline RL, the behaviour policy that is used to collect experience could be completely different from the agent that is currently training based on these experiences \cite{offline-rl-1,offline-rl-2,offline-rl-3}. Cogment handles this through its efficient use of its \textbf{activity logger}.

\subsection{Human-in-the-loop Learning}

\begin{figure}
\label{fig:hitl}
\centering
\includegraphics[width=0.7\columnwidth]{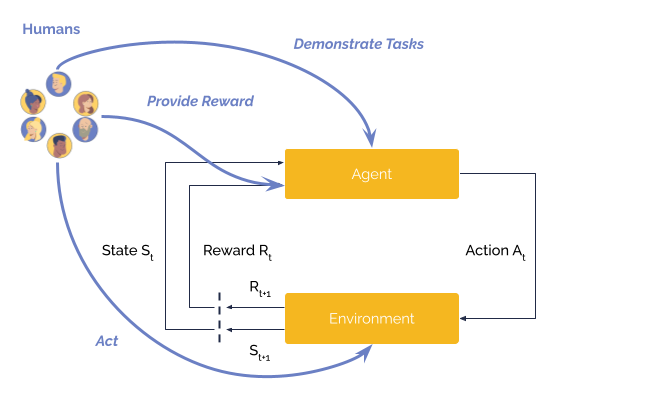}
\caption{An illustration of different ways in which humans can be involved in the training process of an AI agent}
\end{figure}

Humans can interact with learning agents in multiple ways as shown in the Figure \ref{fig:hitl}. They can act in the environment on par with agents (e.g, man vs machine multi-player games), provide rewards to the agents, demonstrate tasks to them, or generate tasks for agents to achieve, and Cogment provides a unifying framework to fulfill all these human-in-the-loop cases. For example, having humans acting in the environment is a way to ensure agents will take safe exploratory actions in sensitive contexts like autonomous driving \cite{human-demonstration-reward-cars,human-demonstration-reward-DM}. 
Humans can provide rewards for several learning algorithms, for example in the context of evaluating machine-generated dialogues \cite{human-feedback-dialogue1,human-feedback-dialogue2,human-feedback-dialogue3}, summaries \cite{summarize-human-feedback-openAI,summarize-human-feedback-2}, semantic parsers \cite{human-feedback-semantic-parser}, natural language \cite{human-feedback-language1}, machine translation \cite{human-feedback-MT} and many others. 

In some cases, it is challenging to design a reward function or, the reward function could be sparse, thus making it hard for an RL agent to learn. In such cases, agents can learn from human demonstrations under the imitation learning (IL) paradigm \cite{IL-survey,GAIL,primalwass}. On classical reinforcement learning tasks such as Cartpole or Mountain Car, the AI Redefined team demonstrated that a combination of RL and IL from human inputs leads to an increase of data efficiency, i.e. require less data for the same performances, when compared to pure RL approaches \cite{air2020}. If it's expensive to use human demonstrations, humans can be used to generate curriculum i.e, generate tasks with an increasing level of difficulty so that the AI agent can learn faster \cite{poet-wang2019paired,human-demonstration-and-curriculum,human-curriculum-DM1}. On the other hand, if no human input is possible, a different AI agent can be used to generate the curriculum \cite{self-play-SukhbaatarKSF17,openai2021asymmetric}.

In section \ref{sec:use-cases}, we explain in detail how we used the Cogment framework for human-in-the-loop learning in two different applications: \emph{Smart Responders} and \emph{Quack Arena}.


\section{Framework}

\subsection{Core Concepts}

Cogment is built around concepts adapted from multi-agent systems (agents, environment), Markov decision processes (action and observation spaces) and reinforcement learning (trials, rewards).

\subsubsection{Trials}

Trials are what a Cogment deployment runs. They enable actors to interact with their environment. Trials are started by clients connecting to Cogment. A trial can end either by being terminated from a client or end by itself, for example once a specific state of the environment is reached.

During the trial:

\begin{itemize}
\item The environment generates \textbf{observations} of its internal state and sends them to the actors who consume them in accordance with their defined observational capabilities.
\item Given these observations, each actor might choose and take an \textbf{action}.
\item The environment receives the actions and updates its state.
\item \textbf{Rewards} can be sent to the actors from either the environment or other actors.
\item Actors receive \textbf{rewards}.
\item The actors or the environment can send \textbf{messages} to actors or the environment.
\item A log of the activity during the trial (observations, actions, rewards \& messages) is produced and can be stored.
\end{itemize}

A trial is defined by the participating actors and the host environment. As a concept, trials are quite close to Reinforcement Learning (RL)'s \textbf{episodes}, i.e. all the states that come between an initial state and a terminal state. However, because Cogment can be used outside of an RL context, we prefer to use the more generic term of trial.

\subsubsection{Actors}

Actors within a trial instantiate actor classes defined by the nature of the information they receive from the environment, their observation space, and what actions they can perform, their action space.

In Cogment, the observation and action spaces are defined as typed data structures. In particular, Cogment uses protobuf \cite{protobuf_v3} as a format to specify these data structures. This typing defines both an interface contract between the actors and the environment and helps convey semantic information, thus facilitating the independent design and development of both.

An actor might be controlled either by a software agent, or by a human. Whichever the case, the process of generating actions based on observations remains the same, and the environment treats them the same.

\subsubsection{Environment}

The environment is the context within which the trial takes place. The environment receives the actions done by the actors, usually updates an internal state, and generates an observation for each actor.

The environment is the main integration point between Cogment and an external system, either a \textbf{simulation} or a \textbf{"real-world" system}.

\subsection{Architecture} \label{sec:architecture}

\begin{figure}
\centering
\includegraphics[width=0.7\columnwidth]{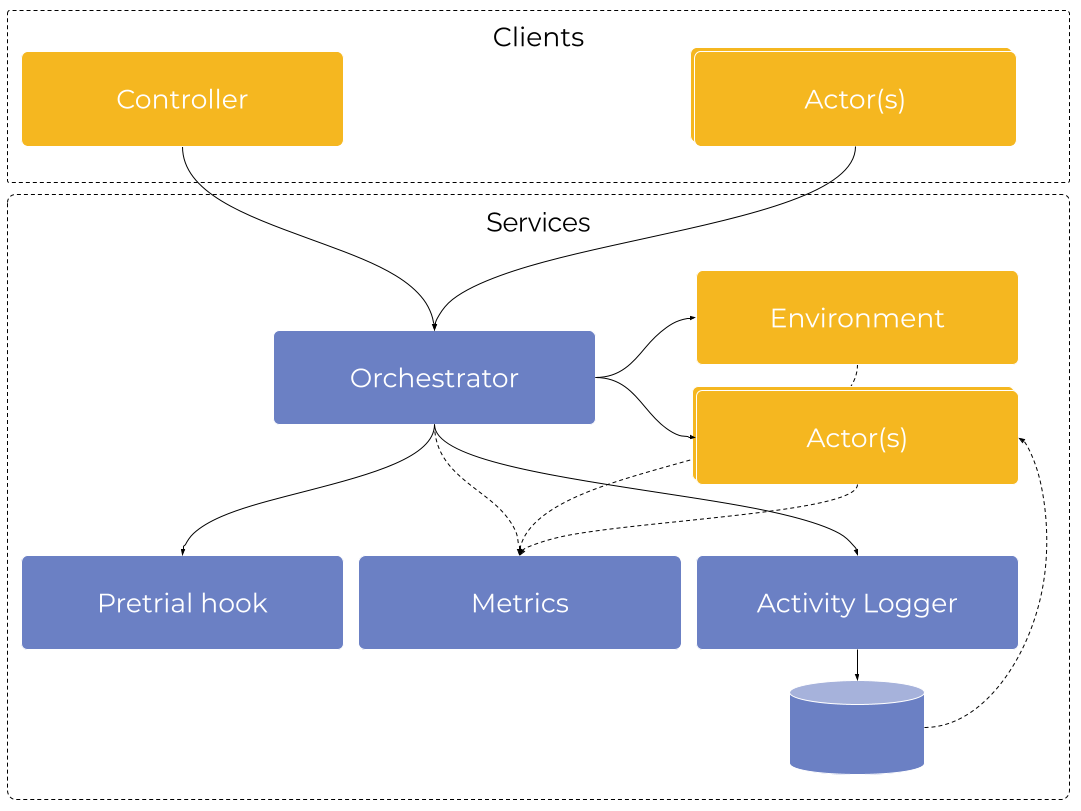}
\caption{Cogment architecture consists of multiple components. These components are either provided by the Cogment framework, depicted above in blue, or implemented for a particular project, depicted above in orange.}
\label{fig:architecture}
\end{figure}

Running trials with Cogment usually involves the deployment of a cluster of services and its clients, cf. figure \ref{fig:architecture}.

Components implemented by users can use one of the Cogment SDKs, or directly implement the underlying protocol (gRPC). Components communicate using gRPC \cite{grpc}, and clients can also communicate in a web-friendly way using gRPC-Web and gRPC-Web proxy \cite{grpcweb}.

\subsubsection{Orchestrator}

The orchestrator is the glue that binds everything together. It is responsible for running the trials and contacting other services as needed to ensure their execution.

The key aspect of Cogment's orchestrator is its capacity to handle a number of network connections in parallel while keeping its responsiveness.

\subsubsection{Controller}

The controller implementation is a key part of using Cogment, it initiates communication with the orchestrator to control the execution of trials. It is responsible for starting trials, retrieving and watching their state (including the end of the trial), or requesting trial termination.

\subsubsection{Environment}

The environment implementation is accessed by the orchestrator to run the environment during trials.

Using one of Cogment's SDKs, the environment can be implemented as a function integrating a \emph{"state of the world"} with the trial. This function performs the following tasks during the Trial:

\begin{itemize}
\item Generate observations from the current \emph{state of the world}, for example retrieving the visible objects from a 3D simulation.
\item Apply the actions, thus updating the \emph{state of the world}, for example changing the velocity of a moving vehicle in a race simulation.
\item Evaluate the performance of actors and send them rewards, for example by checking if a vehicle crossed the finish line in a race simulation.
\item Send and receive direct messages.
\end{itemize}

\subsubsection{Actors}

Actors can be implemented in two different ways, either as a service or as a client.  In both cases, if using a Cogment SDK, the bulk of the code is the same, only the initial setup differs.

\textbf{Service actors} are accessed by the orchestrator during trials, while \textbf{client actors} join a trial by initiating the communication with the orchestrator. Client actor implementations can \emph{reach} a Cogment deployment through NAT traversal \footnote{Network address translation (NAT) traversal is a computer networking technique of establishing and maintaining Internet protocol connections across gateways (sometimes referred to as firewalls) that implement network address translation. cf.~\url{https://en.wikipedia.org/wiki/NAT_traversal}}. This makes them particularly well-suited to implement human-driven actors in any client-side applications, in web-browsers or video game engines for example.

Using one of Cogment's SDKs, actors can be implemented as functions handling the integration between a decision-making actor (software agent or human) and the trial. This function performs the following tasks during the trial:

\begin{itemize}
\item Receive observations and do actions in response, for example vectorizing the retrieved observation, feeding it to a neural network and converting its output to an action.
\item Send and receive rewards, for example using them to update a neural network.
\item Send and receive direct messages.
\end{itemize}

Note that past rewards (and all data) can also be retrieved after-the-fact using an activity logger.

\subsubsection{Additional optional services}

Beyond the core services described above, a Cogment deployment can include these additional ones:

\begin{itemize}
\item \textbf{Pre-trial hooks} can be used to dynamically setup trials from a given configuration, for example changing the number of actors or pointing to other environment or actor implementations.
\item \textbf{Activity logger} can be used to listen to the activity during a trial (actions, observations, rewards, messages), for example to store their data in order to do offline training of AI agents.
\item \textbf{Metrics} centralizes and makes available the history of rewards received by actor implementations as well as computational performances and general metrics of the micro services.
\end{itemize}

\subsection{Other Frameworks and Libraries}

\cite{ikostrikov} is one of the first RL code bases in pytorch that implemented several algorithms like A2C, PPO, ACKTR, GAIL on a wide range of OpenAI gym environments and was successful in replicating the results of these algorithms. However this early example was not optimized for large scale distributed training, a requirement for the training of large scale RL models. 
Concurrently, \cite{rllib} introduced rllib, built on top of distributed machine learning framework Ray \cite{ray}, it provides a lot of in-built algorithms for direct use. rlpyt \cite{rlpyt} is another RL library that provides modular and optimized implementations of several deep RL algorithms in pytorch and is useful for small to medium scale research. Garage \cite{garage} also provides similar utilities and is often used for benchmarking different algorithms. ACME \cite{hoffman2020acme} (that uses reverb \cite{cassirer2021reverb}) is a recent framework and is commonly used for quick prototyping. While most of these frameworks are suitable for RL research, they do not explicitly account for how humans can actively participate in the learning process.

Designed for multi-agent systems research, MAS, frameworks such as Jade \cite{bellifemine1999jade}, Spade \cite{gregori2006jabber} or Mesa \cite{masad2015mesa} are very well suited for implementing and deploying autonomous agents. They also usually provide ways for humans to impersonate an agent in the system. However they are not designed to be integrated with complex simulated environment. Because of their decentralized nature, retrieving the data stream generated by the agents is either difficult or inefficient, making them ill suited for machine learning, especially online machine learning. These limitations explains why MAS framework, while designed for multi-agent are not usable for multi agent reinforcement Learning (MARL) or HITL Machine Learning.


\section{Use cases} \label{sec:use-cases}

In this section, we present two use cases demonstrating how multi-agent and human-in-the-loop settings can be implemented using Cogment.

\subsection{Quack Arena}

Quack Arena was developed as a testbed and showcase for multi agent reinforcement learning (MARL) using Cogment. The game is a competitive and cooperative paintball-like shooter, where teams of agents compete against each other in an arena. Agents shoot paint balls at their opponents from other teams. An agent is eliminated if it is hit by an opponent; there is no friendly fire. Last team standing wins the game; each Cogment's trial consists of a game. 

\begin{figure}[htp]
    \centering
    \includegraphics[width=0.7\textwidth]{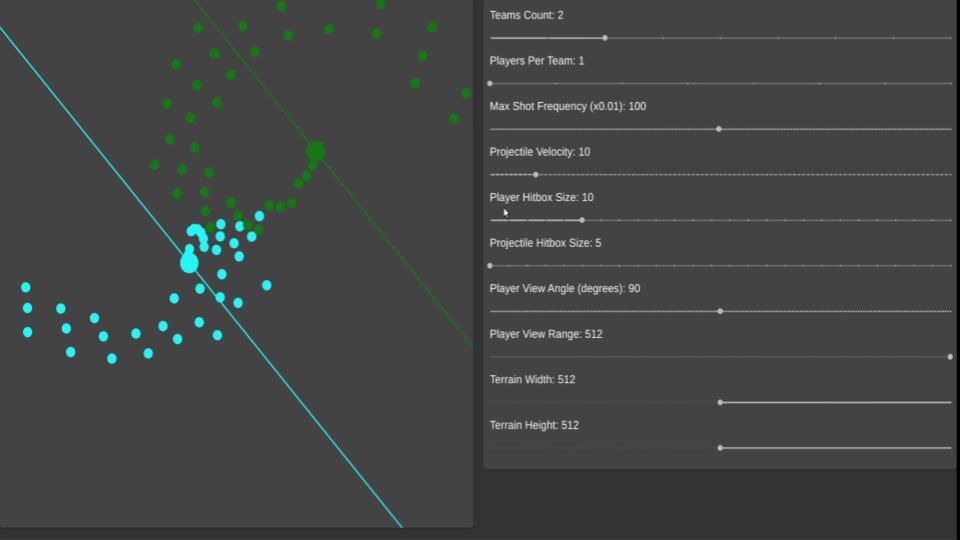}
    \caption{Screenshot of the Quack Arena Web Client. The left pane of the window shows the arena during a trial, while the right pane displays the configurations of the trial such as the number of teams in the trial, number of players in each team, shot frequency, shot velocity, arena size etc.}
    \label{fig:Quackarena-arena}
\end{figure}

Quack Arena includes a lightweight web based front-end where users can configure and watch a trial. Figure \ref{fig:Quackarena-arena} is a screenshot of this Web Client featuring a duel between two trained agents. The larger circles are the players, the smaller circles are the paint balls, and the lines represent the region of visibility for each player. Since the velocity of the paint balls is low, the physics of the simulation makes their trajectory nonlinear with the movements of the players.

Quack Arena also allows to run in a "headless" mode, i.e. without any visualization, to gather data and train multiple agents over campaigns of multiple trials executed in parallel and at scale.

\subsubsection{Modeling}

Quack Arena actually involves two types of agents, the players involved in the game, as well as an observer which purpose is to feed the web client visualization for a particular game. We will focus on the former in the following. All the player agents are implemented as Cogment actors of the same class, they all share the same observation space and action space, and receive rewards in the same way.

\paragraph{Observation Space}

The observation of each agent consists of the position and orientation of itself and all other visible players and objects. In addition, every agent knows its own and other visible players dead/alive status. Self-position (i.e $\{x,y\}$ coordinates) and orientation are relative to the arena's frame of reference, whereas the position and orientation of other visible players and objects are relative to self. 

\paragraph{Action Space}

Actions include weapon's firing, left/right movement, forward/backward movement, and rotation. Each of the movement's values range between $[-1, 1]$ with negative values referring to left, backward and anti-clockwise movements and positive values relating to right, forward and clockwise movements respectively.

\paragraph{Rewards}

The agents receive a reward from the environment for eliminating opponents and a penalty for being terminated but also for wasting time, to avoid having agents learning not to engage to limit the risk of a termination. In the case of one player per team, rewards are bounded between $(-2, 1]$ for each player:

\begin{center}

$\text{time waste penalty} = \frac{-1}{\text{max trial length}}$ 

$\text{termination penalty} = -1$

$\text{on-target reward} = \frac{1}{\text{no.of opponents}}$
\end{center}

\subsubsection{Implementation} \label{sec:Quackarena-implementation}

\begin{figure}[htp]
    \centering
    \includegraphics[width=0.7\textwidth]{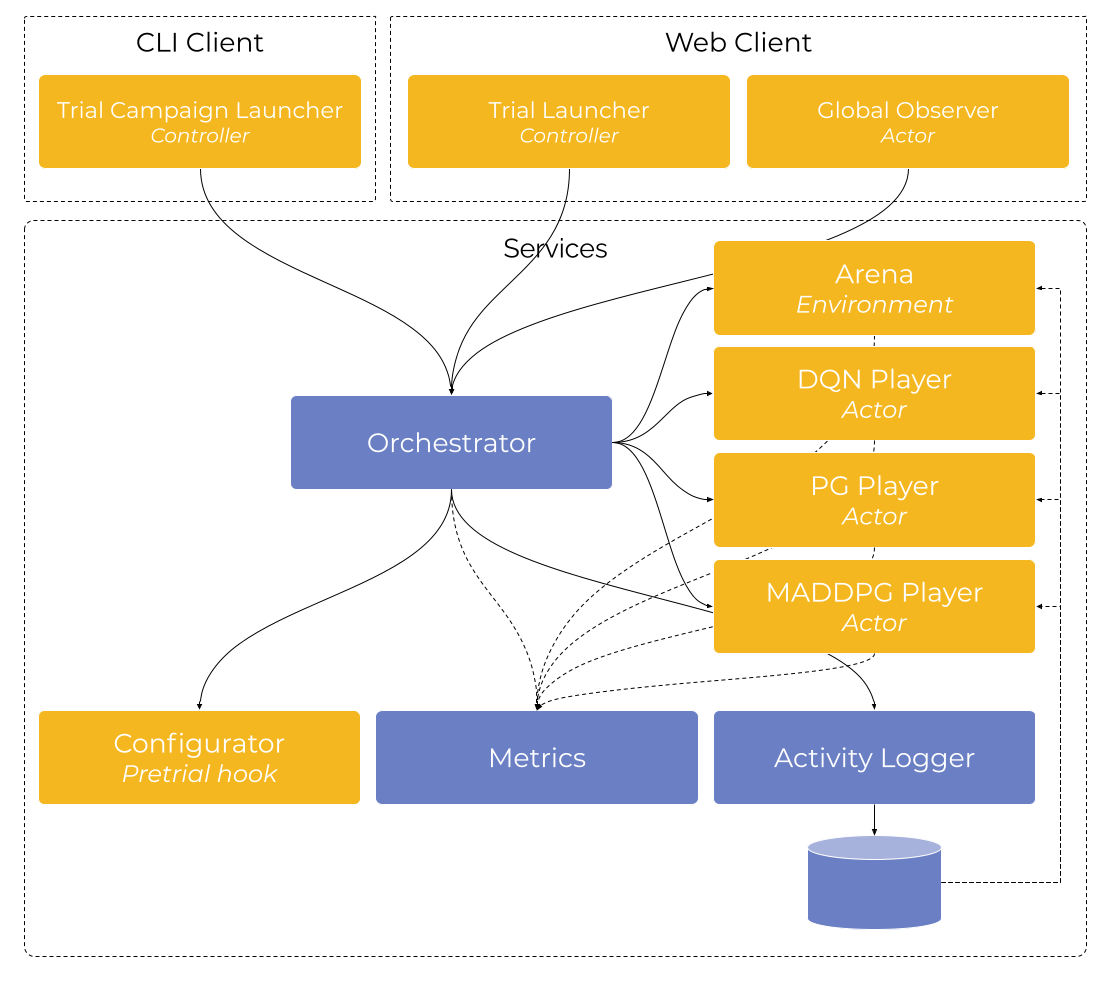}
    \caption{Quack Arena instantiates Cogment Architecture depicted in Figure \ref{fig:architecture} implementing a command line interface (CLI) client including a trial campaign launcher, a web client including a trial launcher as well as an observer actor, and finally a number of services, the arena environment, a trial configurator and a number of Actors. All of the Quack Arena specific modules are depicted in orange.}
    \label{fig:Quackarena-architecture}
\end{figure}

Quack Arena architecture is depicted in Figure \ref{fig:Quackarena-architecture}. Several modules were developed specifically for Quack Arena using the Cogment SDKs.  
\begin{itemize} 
\item The \textbf{arena} is the environment service, it implements the logic for the Quack Arena game: terrain, physics "simulation", players elimination, and reward function
\item The \textbf{web client}, shown in Figure \ref{fig:Quackarena-arena}, allows to configure trial parameters through a controller and and visualize the trial in progress thanks to a global observer actor.
\item The \textbf{command line interface (CLI) client} triggers a number of parallel trials continuously for training purposes.
\item Several \textbf{player agents} are implemented as actor services. Several RL algorithms are available, some able to leverage the continuous nature of the action space, such as MADDPG (Multi Agent Deep Deterministic Policy Gradients) \cite{DBLP:journals/corr/LoweWTHAM17}, others needing to discretize it such as REINFORCE \cite{sutton:nips12, zhang2020sample} or DQN \cite{atari}. Other non-learning agent were implemented using scripted heuristic behaviors, which can be used during training or as an evaluation baseline.
\item Finally the \textbf{configurator} is a pre-trial hook for service discovery and load balancing. Upon startup, actor services register with the configurator. When a trial starts, the configurator can then set the trial's configuration with a random network endpoint of the desired actor implementation to enable the orchestrator to reach them and spread the load between several services.
\end{itemize}

To understand how all the pieces work together, let's look at how a trial plays out. First, one of the controllers defines a basic trial configuration, this includes the number of teams, which player agent should be part of each team, as well as the configuration parameters for the arena such as the speed of the paint balls. When the trial is started by the web client's controller, it adds the global observer actor to the configuration. Once the configuration is complete, the start of the trial is requested to the orchestrator.

As soon as the orchestrator receives the trial start request it calls the configurator pre-trial hook which extends the trial configuration with the network endpoints of the actor implementations involved in the trial. If the global observer client actor is involved, the orchestrator waits for it to join, and the trial can start.

Every trial begins with the arena environment sending initial observations to every actor in the trial through the orchestrator. After receiving observations from the environment, the actor implementations decide on actions according to their respective model. Actions are then sent to the environment through the orchestrator. Once received, it updates its internal state accordingly (e.g. it moves players that decided on a movement, eliminate players hit by a paint ball), it then compute the reward for each player and then sends new observations and rewards to their respective actors. During the trial (or only at the end, according to settings), the orchestrator dumps all observations, actions, rewards and messages exchanged to the activity logger.  The actors can sample from this log to update their respective deep learning models and learn from their experiences so far. The total rewards the actors receive across trials can be monitored through the metrics module's dashboard, the actors' models are considered trained if the moving average of the total rewards they accumulated across ten consecutive trials is more than a predefined threshold.

\subsubsection{Results}

During the experimentation, we observed that all the agents converged to extremely high scores, and demonstrated varying strategies to target opponents and dodge paint balls. Some also learnt to exploit unexpected properties of the environment to gain advantage against their opponents, such as positioning themselves at a corner of the arena where paint-balls from opponents would not hit them. Training was conducted with several topologies from one-versus-one games to melees of heterogeneous teams, with similar degrees of success. From a software and architecture point of view, Quack Arena enabled us to validate the scalability of Cogment. In particular we demonstrated how Cogment facilitates the execution of a massive number of heterogeneous trials in parallel, distributed over several nodes. The Quack Arena project is actively used to continuously test new Cogment releases as well as new agent architecture and training strategies.

\subsection{Smart Responders}

Emergency number calls (like 911) are handled by teams of emergency operators in dispatch centers. Their job is to assess incidents and match resources (e.g. ambulance, firefighting truck) to those emergencies. Emergency operators are often overwhelmed by the number of calls they receive during their shifts, while contending with the complexity inherent in resource allocation, which in turn leads to delays in attending to some emergencies. \textbf{Smart Responders} was developed to demonstrate how a collaboration between humans and AIs could improve the overall operational efficiency of the system by automating the attendance to low priority events while freeing up cognitive space for the operators to tackle higher priority ones or more demanding sub-tasks (typically the callers themselves and the assessment of a situation). While agents automate administrative tasks, provide recommendations on the best units to dispatch to events with regards to their ETA, and handle dispatches to minor incidents for human operators, human operators focus on attending to major events as well as the human element of the work.

\begin{figure}[htp]
    \centering
    \includegraphics[width=\textwidth]{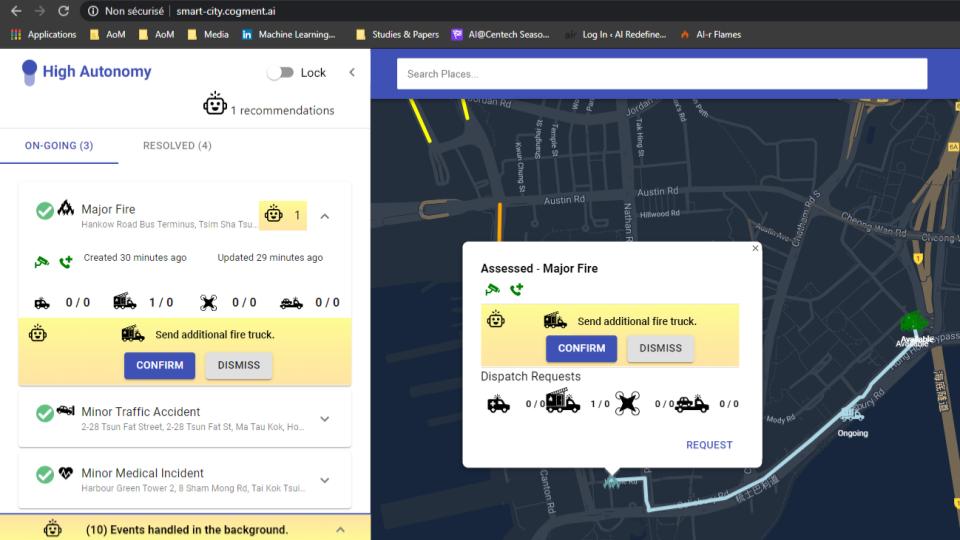}
    \caption{The operator’s web client. On the right is a map of the city where events pop up, and on the left is a list of those events, sorted by priority, and dispatch actions that the recommender agent recommends to the operator. The operator could accept/override either from the list view or by clicking on the events individually from the map.}
    \label{fig:smart-response-webclient}
\end{figure}

The setup consists of a recommender agent that learns to recommend dispatch actions to the human operator, the non-learning dispatch agents, such as the fire trucks, ambulances, tow trucks, and survey drones which service emergencies on the ground, and the human operator who accepts/overrides the recommendations of the recommender. The operator has a birds-eye view of the city or area he or she is in charge of through a web client, as depicted on the right side of the Figure \ref{fig:smart-response-webclient}, and is made aware of emergency incidents of various types and severity levels as they pop up on the map along with their respective dispatch recommendations by the recommender,  as shown in the left panel in Figure \ref{fig:smart-response-webclient}. The operator can either accept the recommender's dispatch suggestions by clicking on the "Confirm" button on the recommendations, or manually dispatch units to events through the events list or the city map. In addition, the operator can toggle the recommender's "Autonomy level", displayed in the top left corner of Figure \ref{fig:smart-response-webclient}, between "High" and "Low" so as to govern the amount and severity of events that the recommender has autonomy over. The recommender agent is able, unless locked out of it by the operator, to toggle itself from high to low autonomy and vice-versa. The autonomy level controls, and recommendation approval features of the system ensure that the human operator has complete authority over the recommender, and control over the operations of the system at all times. 

\subsubsection{Modeling}

The Smart Responders project involves three types of actors.

\paragraph{Dispatch units} Like in the real world, dispatch units are independent actors in Smart response. The dispatch units primary task is to reach an event location as directed either by the human responder or by the recommender, this is conveyed through their observation space consisting of a list of way points. Their action space includes their speed and orientation, as well as their availability status. The latter informs the human responder and the recommender whether they are currently available, on-their-way, or busy.

\paragraph{Recommender} The recommender is the AI agent assisting the human responder. Its observation space consists of its level of autonomy (and whether it has been locked in place by the human operator), the list of emergency events with their location, type, severity, and status, as well as the status and locations of the dispatch units. The recommender's action space consists of a list of recommendations, mapping events with dispatch units. Events can be of type "Fire", "Traffic Accident", or "Medical", and can be of "Major" or "Minor" severity. While the status of an event reflects whether it is ongoing or has been resolved, the status of the dispatch units informs the recommender’s if they have already been assigned to an event.

\paragraph{911 Responder} The 911 responder is the human operating the dashboard. The primary responsibility of the 911 responder within the tool is to approve or override dispatch recommendations from the recommender, or dispatch units manually to those events when deemed appropriate. Their observation space includes the locations of ongoing events and available dispatch units, dispatch recommendations from the recommender, and the state of the "Autonomy Level". The action space of the 911 responder includes toggling the recommender's "Autonomy Level" between "High" and "Low", "Locking/Releasing" its ability to change its "Autonomy Level" on its own, "Confirming/Dismissing" its dispatch recommendations, and dispatching units to events manually. 

\subsubsection{Implementation}

\begin{figure}[htp]
    \centering
    \includegraphics[width=0.7\textwidth]{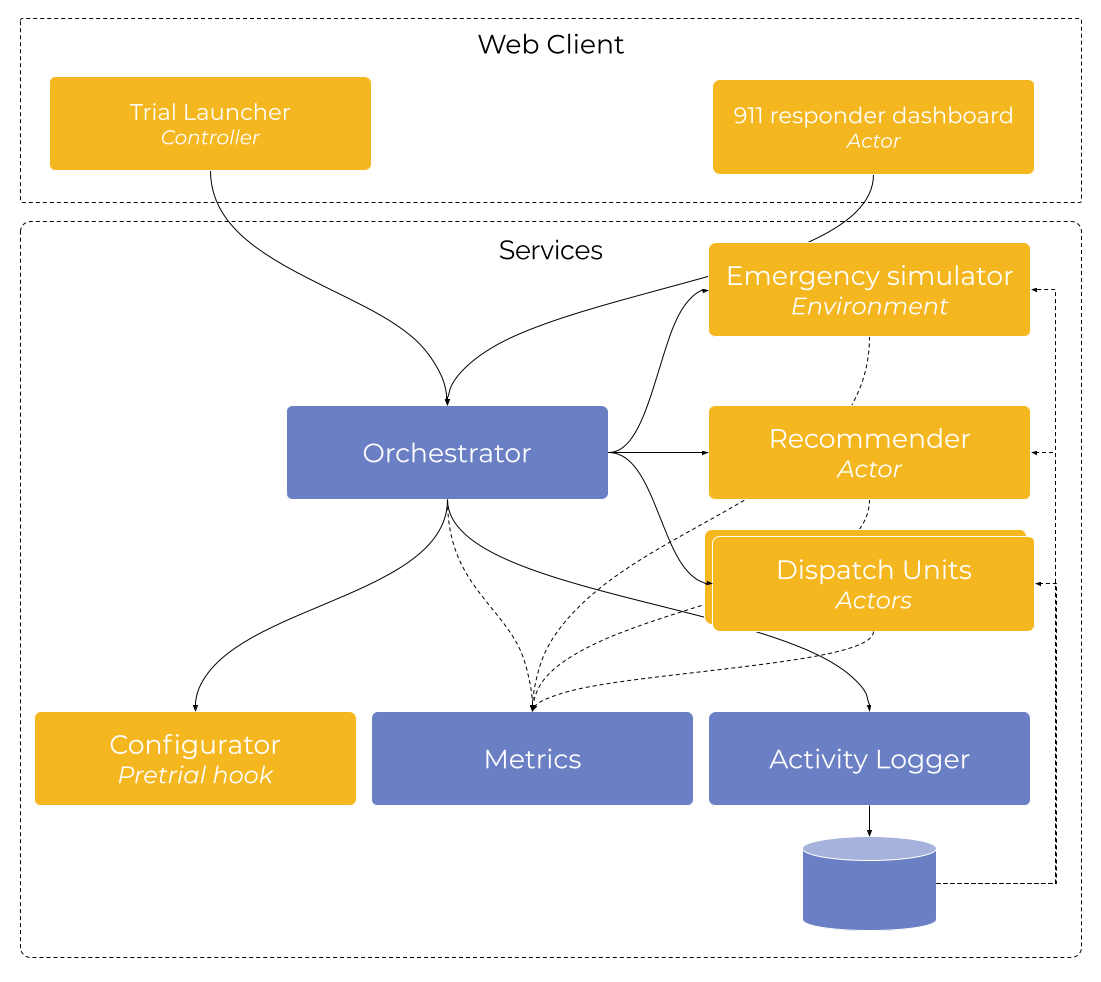}
    \caption{Smart Responders instantiates Cogment Architecture depicted in Figure \ref{fig:architecture} implementing a web client including a trial launcher as well as an actor for the human 911 responder, and finally a number of services, the simulator, a trial configurator, the recommender actor and a number of dispatch units actors. All of the Smart Responders specific modules are depicted in orange.}
    \label{fig:smart-response-architecture}
\end{figure}

Smart Responders architecture is depicted in Figure \ref{fig:smart-response-architecture}. The architecture is quite similar to Quack Arena, described in section \ref{sec:Quackarena-implementation}, thus we will focus on describing the main modules that are different.
\begin{itemize} 
\item The \textbf{web client} also shown in Figure \ref{fig:smart-response-webclient}, allows for humans to interact with agents (and configure trial parameters prior to their running). 
\item The \textbf{recommender} actor implementation allocates units to events given its "Autonomy Level", using standard resource allocation optimizers.
\item The \textbf{dispatch units} actor implementations are non-learning in nature, and implement classical path following algorithms from the way points received as observations. They follow those way points by increasing/decreasing their speed and changing orientation as required.
\end{itemize}

Trial execution is again quite similar to Quack Arena described in section \ref{sec:Quackarena-implementation}. The main difference is that the client actor is not only an observer; it takes an active part in the trial. Furthermore, in order to measure the effectiveness of human-AI teaming on the operational efficiency of the system, the controller can launch trials in three different modes, "human only", "AI only" and "human-AI". In the "human only" mode, the responder doesn't receive any assistance from the recommender, whereas in the "AI Only" mode, the recommender operates with complete autonomy without a human in the loop. The "human-AI" mode, on the other hand, requires the human operator and the recommender to collaborate as efficiently as possible so as to outperform other modes of operations. 

Every trial begins with a few dispatch units and events of different types and severity distributed across the map stochastically. After receiving observations from the environment, the recommender sends dispatch recommendations to the human responder who, over the course of trials, learns the best strategy to accept/override recommendations as well as set the "Autonomy Level". After receiving dispatch directives, either from the the recommender or the human operator, the dispatch unit services send their movement actions to the environment that updates the environment's state and sends the updated observations to all the actors. The environment also creates emergency events during the course of the trial.

\subsubsection{Results}

By analyzing the data collected over a trial campaign, we observed that "human-AI" teaming out performed "human-only" and "AI-only" settings in scenarios where the frequency of event generation was high and the underlying process that generated the events was non-stationary. These are scenarios that resemble the real world more closely. While humans were better at performing under uncertainty, the recommender complemented the humans by selecting units and serving events at a higher frequency.

\section{Conclusion}

Guidance from humans, for example in the form of evaluation or demonstration, is key to design, train, and deploy a new generation of AI. Cogment provides a unifying framework for operating and training AI agents in simulated or real-world environment in collaboration with humans. It has been designed to scale from small prototypes running on a personal workstation, to training campaign running on distributed clusters to continuous operations integrated in production systems while being easy to use for researchers, developers and infrastructure people alike. By making it open source, we hope it will drastically increase the pace of research and development and deployment of several human-in-the-loop or reinforcement learning applications in the real world.  

\section*{Acknowledgements}

Cogment was developed from contribution of the AI Redefined engineering team including Sara Brazille, François Chabot, Jonathan M. Fisher, Sai Krishna Gottipati, Sagar Kurandwad, Clodéric Mars, Yves Paradis and Vincent Robert.

\printbibliography

\end{document}